\documentclass{article}

\usepackage{microtype}
\usepackage{graphicx}
\usepackage{subfigure}
\usepackage{booktabs} 

\usepackage{hyperref}

\usepackage[accepted]{icml2025}

\usepackage{amsmath}
\usepackage{amssymb}
\usepackage{mathtools}
\usepackage{amsthm}
\usepackage{multirow}
\usepackage{multicol}

\usepackage[capitalize,noabbrev]{cleveref}

\theoremstyle{plain}

\theoremstyle{definition}

\theoremstyle{remark}

\usepackage[textsize=tiny]{todonotes}

\icmltitlerunning{Representation Learning of 
Event Time Series with Sparse Autoencoders}

\begin{document}

\twocolumn[
\icmltitle{Learning Representations of Event Time Series with Sparse Autoencoders for Anomaly Detection, Similarity Search, and Unsupervised Classification}



\icmlsetsymbol{equal}{*}

\begin{icmlauthorlist}
\icmlauthor{Steven Dillmann}{a}
\icmlauthor{Juan Rafael Martínez-Galarza}{b}

\end{icmlauthorlist}

\icmlaffiliation{a}{Stanford University}
\icmlaffiliation{b}{Center for Astrophysics $|$ Harvard \& Smithsonian}

\icmlcorrespondingauthor{Steven Dillmann}{stevendi@stanford.edu}

\icmlkeywords{Machine Learning, ICML}

\vskip 0.3in]



\printAffiliationsAndNotice{}  

\begin{abstract}
Event time series are sequences of discrete events occurring at irregular time intervals, each associated with a domain-specific observational modality. They are common in domains such as high-energy astrophysics, computational social science, cybersecurity, finance, healthcare, neuroscience, and seismology. Their unstructured and irregular structure poses significant challenges for extracting meaningful patterns and identifying salient phenomena using conventional techniques. We propose novel two- and three-dimensional tensor representations for event time series, coupled with sparse autoencoders that learn physically meaningful latent representations. These embeddings support a variety of downstream tasks, including anomaly detection, similarity-based retrieval, semantic clustering, and unsupervised classification. We demonstrate our approach on a real-world dataset from X-ray astronomy, showing that these representations successfully capture temporal and spectral signatures and isolate diverse classes of X-ray transients. Our framework offers a flexible, scalable, and generalizable solution for analyzing complex, irregular event time series across scientific and industrial domains.
\end{abstract}

\section{Introduction}

Event time series—irregular sequences of timestamped discrete events of a particular observational modality—arise across a wide range of scientific and industrial domains. In astronomy, they represent photon arrivals detected by telescopes; in computational social science, they trace social media activity or user interactions; in cybersecurity, they log system alerts; in finance, they capture transactions and market fluctuations; in healthcare, they reflect patient monitoring data; in neuroscience, they record neuronal spike trains; and in seismology, they track seismic wave arrivals. These time series encode rich temporal and domain-specific information, often containing salient patterns, natural groupings, and rare anomalies. Yet this information can remain inaccessible, as the irregular and unstructured nature of event time series limits the effectiveness of conventional analysis techniques. Unlocking their latent structure requires more flexible and expressive methods that simultaneously enable a range of follow-on analyses. In this work, we present a representation learning \cite{bengio2013representation} framework based on sparse autoencoders (SAEs; \citealt{hinton2006reducing, ng2011sparse}) to extract meaningful features from event time series, enabling downstream tasks such as anomaly detection, similarity searches, clustering, and unsupervised classification. By enforcing sparsity in the latent space, the model is encouraged to focus on the most physically relevant features, while remaining robust to nuisance variation arising from noise, observational systematics, or contextual artifacts unrelated to the underlying phenomena. 

We demonstrate our pipeline on archival data from the Chandra X-ray Observatory \cite{2000SPIE.4012....2W}. In X-ray astronomy, event files record the arrival times and energies of individual photons, capturing the stochastic behavior of astrophysical sources. Some of the most interesting discoveries—such as extragalactic fast X-ray transients (FXTs; \citealt{2013ApJ...779...14J, 2015MNRAS.450.3765G, 2019Natur.568..198X, 2022ApJ...927..211L}) and even the first candidate extragalactic planet \citep{2021NatAs...5.1297D}—have been made serendipitously, often years after the data were collected. FXTs are brief, intense X-ray flares from extragalactic sources lasting only minutes to hours, whereas the planet candidate was identified as a dip in the X-ray light curve. These and other transient sources span a wide range of timescales and spectral properties, offering insights into their astrophysical origins. However, conventional methods often rely on rigid assumptions and handcrafted summary statistics \cite{2019MNRAS.487.4721Y,quirola2022extragalactic}, limiting their ability to capture the full complexity of such phenomena. In contrast, our sparse representation learning approach provides a flexible and scalable alternative—well-suited for anomaly detection, transient searches, and unsupervised classification in large datasets containing diverse types of event time series. Our method has already led to the discovery of XRT 200515 \cite{chandra_representation}, a previously unknown FXT and a new hyperluminous supersoft X-ray source \cite{sacchi2025} that had been buried in the archive for years.




\section{Method}

\subsection{Tensor Representation of Event Time Series} \label{sec:tensor}

For a dataset with $n$ event time series $\{\boldsymbol{x}_i\}^n_{i=1}$ that vary in length $N$ and duration $T$, we introduce a fixed-size tensor representation that standardizes the data for subsequent analysis. This approach is inspired by the DMDT maps developed for optical light curves by \citet{2017arXiv170906257M}. 

Assume that each event time series consists of a sequence of timestamps $\boldsymbol{t} = \{t_k\}_{k=1}^N$ and associated event modality values $\boldsymbol{E} = \{E_k\}_{k=1}^N$. In the context of X-ray astronomy, $\boldsymbol{t}$ would correspond to the photon arrival times and $\boldsymbol{E}$ to the photon energies. The total duration of the event time series is given by $T = t_N - t_1$, and the event values lie in the range $[E_{\min}, E_{\max}]$, where $E_{\min} = \min_k E_k$ and $E_{\max} = \max_k E_k$. To standardize the inputs across event time series, we normalize the time axis as
\begin{equation}
\boldsymbol{\tau} = \frac{\boldsymbol{t} - t_1}{T},
\end{equation}
which maps all timestamps to the normalized interval $\boldsymbol{\tau} \in [0, 1]$. For the event values, we apply a transformation
\begin{equation}
\boldsymbol{\epsilon} = f(\boldsymbol{E}),
\end{equation}
with $\boldsymbol{\epsilon} \in [\epsilon_{\min}, \epsilon_{\max}]$,
where $\epsilon_{\min} = f(E_{\min})$ and $\epsilon_{\max} = f(E_{\max})$. The choice of transformation function $f(\cdot)$ depends on the domain and distribution of the event modality and is intended to highlight relevant structure for downstream analysis. To capture local temporal dynamics beyond absolute event times, we introduce a third dimension based on inter-event intervals. Specifically, we compute the time differences between consecutive events, $\boldsymbol{\Delta t} = \{\Delta t_k\}_{k=1}^{N} = \{t_{k+1} - t_k\}_{k=1}^{N-1}$, and normalize them to the range $[0, 1]$ as
\begin{equation}
\boldsymbol{\Delta\tau} = \frac{\boldsymbol{\Delta t} - \Delta t_{\min}}{\Delta t_{\max} - \Delta t_{\min}},
\end{equation}
where $\Delta t_{\min}= \min_k \Delta t_k$ and $\Delta t_{\max}= \max_k t_k$. This third axis captures local temporal density and is a proxy for the event rate: small values of $\Delta\tau$ correspond to periods of concentrated activity, while larger values indicate sparser or quiescent intervals. In heterogeneous, irregular discrete time series, temporal spacing often conveys essential structural information beyond the absolute timestamps alone. Including this dimension improves the expressiveness of the representation, allowing models to capture transient dynamics and distinguish between different temporal regimes. 

These three normalized axes—event time ($\boldsymbol{\tau}$), transformed modality ($\boldsymbol{\epsilon}$), and inter-event interval ($\boldsymbol{\Delta\tau}$)—are discretized into fixed-width bins with $n_{\tau}$, $n_{\epsilon}$, and $n_{\Delta\tau}$ bins, respectively. Each event is assigned to a voxel in the resulting three-dimensional histogram based on its coordinates in this normalized space. The number of bins along each axis can be chosen flexibly to achieve the desired level of temporal or modality-specific resolution, depending on the application, domain, and characteristics of the data. This allows the representation to be tailored to the appropriate scale and granularity for a given task, while still providing a consistent input format for models, even when the dataset contains sequences of variable length and irregular temporal structure.
This results in a tensor $\boldsymbol{X}_i\in \mathbb{R}^{n_\tau \times n_\epsilon \times n_{\Delta\tau}}$, which provides a standardized, structured input format for each of the original event time series $\boldsymbol{x}_i$. Each tensor is then collected into a final processed dataset $\{\boldsymbol{X}_i\}^n_{i=1}$. We also refer to these tensors as $E$–$t$–$dt$ cubes. If we were to only bin along time $\boldsymbol{\tau}$ and modality $\boldsymbol{\epsilon}$, we would obtain two-dimensional tensors $\boldsymbol{X}_i\in \mathbb{R}^{n_\tau \times n_\epsilon}$, which we refer to as $E$–$t$ maps. Appendix \ref{appendix1} shows $E$–$t$ maps and $E$–$t$–$dt$ cubes for an X-ray flare, dip and pulsating source.

\begin{figure*}[t]  
    \centering    \includegraphics[width=1\linewidth]{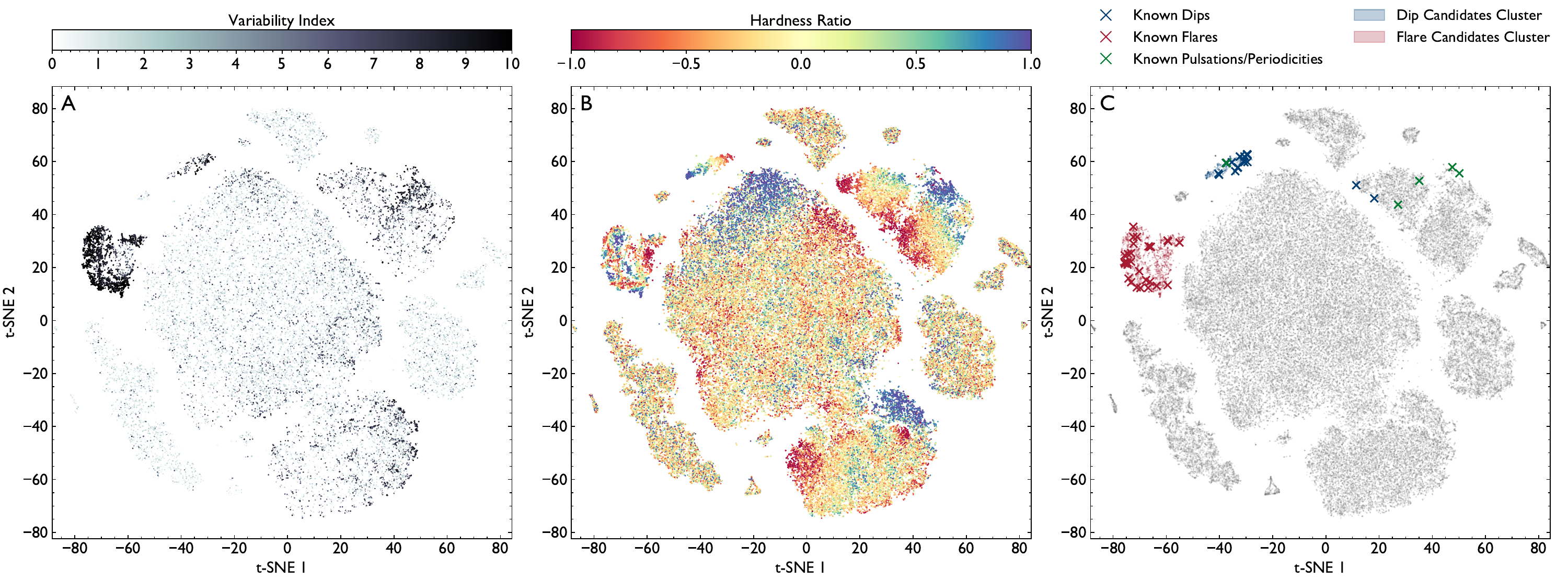}
    \caption{Two-dimensional t-SNE projection of the learned latent space from the SAE applied on the $E$–$t$-$dt$ cubes. Panel A: Points are color-coded by the variability index of the corresponding X-ray sources. Panel B: Points are color-coded by the hard-to-soft X-ray hardness ratio. Panel C: Known dips, flares, and pulsating sources (crosses) occupy distinct clusters in the embedding space, enabling the identification of new transient candidates via clustering and similarity searches.}
    \label{fig:tsne}
\end{figure*}

\subsection{Sparse Representation Learning}

We now aim to learn compact and informative representations from the $E$–$t$–$dt$ cubes. To this end, we employ a SAE—an unsupervised neural network model designed to extract structured, low-dimensional features from high-dimensional inputs. By learning to reconstruct the input data while enforcing sparsity in the latent space, the SAE identifies salient patterns and compresses each tensor into a representation that preserves essential temporal and modality-specific features, while remaining insensitive to irrelevant variation \cite{ranzato2007sparse, ng2011sparse}. This includes robustness to noise, systematics, observational artifacts, and context-dependent features introduced by the data acquisition process. Instead, the learned representations are encouraged to capture the most important latent factors that reflect the underlying physical characteristics and meaningful signals encoded in the time series, rather than superficial or incidental aspects of the data. 

Formally, let $\boldsymbol{X}_i\in \mathbb{R}^{n_\tau \times n_\epsilon \times n_{\Delta\tau}}$ denote the input tensor for the $i$-th event time series. The autoencoder consists of two parameterized functions: (i) an encoder network $\phi_\theta: \mathbb{R}^{n_{\tau} \times n_{\epsilon} \times n_{\Delta\tau}} \rightarrow \mathbb{R}^d$ that maps $\boldsymbol{X}_i$ to a $d$-dimensional latent vector $\boldsymbol{z}_i = \phi_\theta(\boldsymbol{X}_i)$, and (ii) a decoder network $\psi_\theta: \mathbb{R}^d \rightarrow \mathbb{R}^{n_{\tau} \times n_{\epsilon} \times n_{\Delta\tau}}$ that reconstructs the input as $\hat{\boldsymbol{X}}_i = \psi_\theta(\boldsymbol{z}_i)$. To encourage sparsity in the learned representations, we impose an L1 penalty on the latent vector $\boldsymbol{z}_i$. The total training loss balances reconstruction accuracy and sparsity, and is given by:
\begin{equation}
\mathcal{L}(\theta) = \frac{1}{n} \sum_{i=1}^{n} \left\| \boldsymbol{X}_i - \hat{\boldsymbol{X}}_i \right\|_2^2 + \lambda \left\| \boldsymbol{z}_i \right\|_1,
\end{equation}
where $\lambda > 0$ controls the strength of the sparsity regularization. Once trained, we discard the decoder and use the encoder $\phi_\theta$ as a standalone feature extractor. The resulting latent vectors $\boldsymbol{z}_i$ serve as learned representations of the original event time series and can be directly used for downstream tasks such as clustering, anomaly detection, or similarity-based retrieval. A similar procedure is followed when using the $E$–$t$ maps as inputs to a SAE.

\section{Experiments on Downstream Tasks}

\paragraph{Application to X-ray Astronomy} 

We apply our method to data from the Chandra Source Catalog (CSC) version 2.1 \citep{2024arXiv240710799E}. Specifically, we use region event files, which list individual photon events for a specific source in a given observation—each with associated arrival time and energy. The dataset comprises 95,473 event files from 58,932 unique X-ray sources. For further details on the dataset, please refer to \citealt{chandra_representation}. Using the terminology introduced in Section \ref{sec:tensor}, the photon arrival times correspond to $\boldsymbol{t}$ and the photon energies to $\boldsymbol{E}$. We choose the event transformation function $f(\cdot)$ to be the base-10 logarithm, i.e., $f(\boldsymbol{E}) = \log_{10}(\boldsymbol{E})$, which is commonly used in X-ray astronomy. For the $E$–$t$–$dt$ cubes, we choose a dimensionality of $(n_{\tau}, n_{\epsilon}, n_{\Delta \tau})=(24, 16, 16)$. For the SAE, 
we choose a sparsity strength of $\lambda = 0.1$ and its latent vector dimension is 24. We also run the experiments described in this section on the $E$–$t$ maps with a dimensionality of $(n_{\tau}, n_{\epsilon})=(24, 16)$. In this case, we use a SAE with convolution layers and a latent vector dimension of 12. More details are provided in Appendix \ref{appendix2}. 

\begin{figure*}[t]  
    \centering
\includegraphics[width=0.245\linewidth]{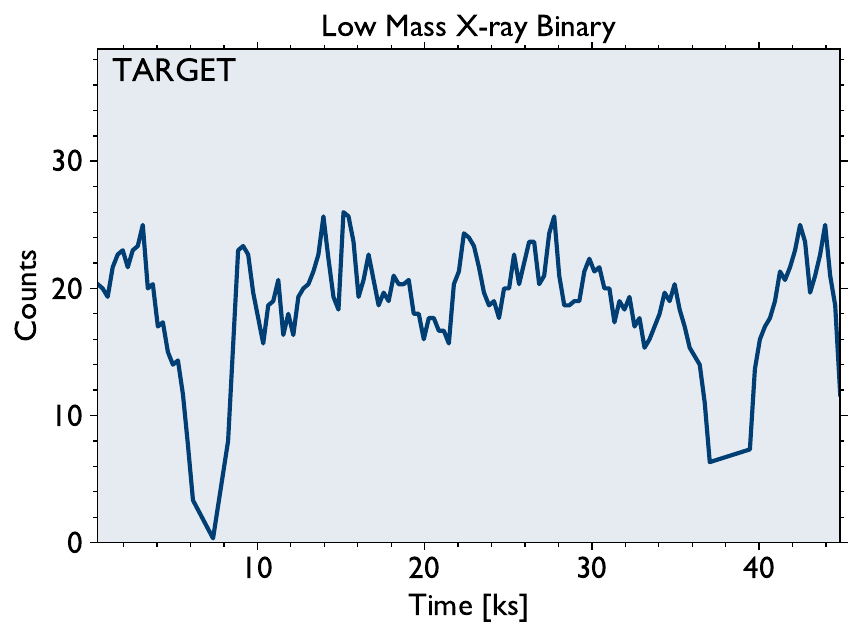}
\includegraphics[width=0.245\linewidth]{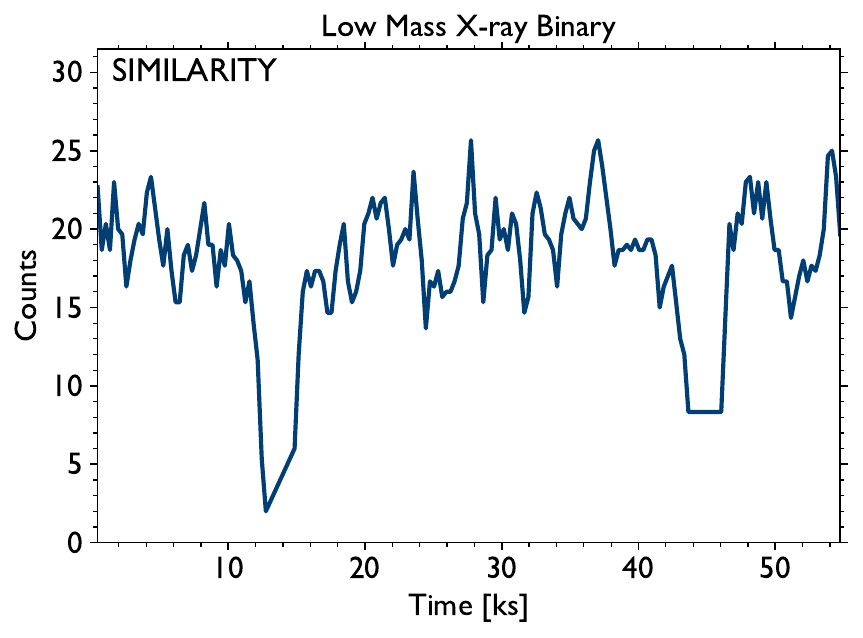}
\includegraphics[width=0.245\linewidth]{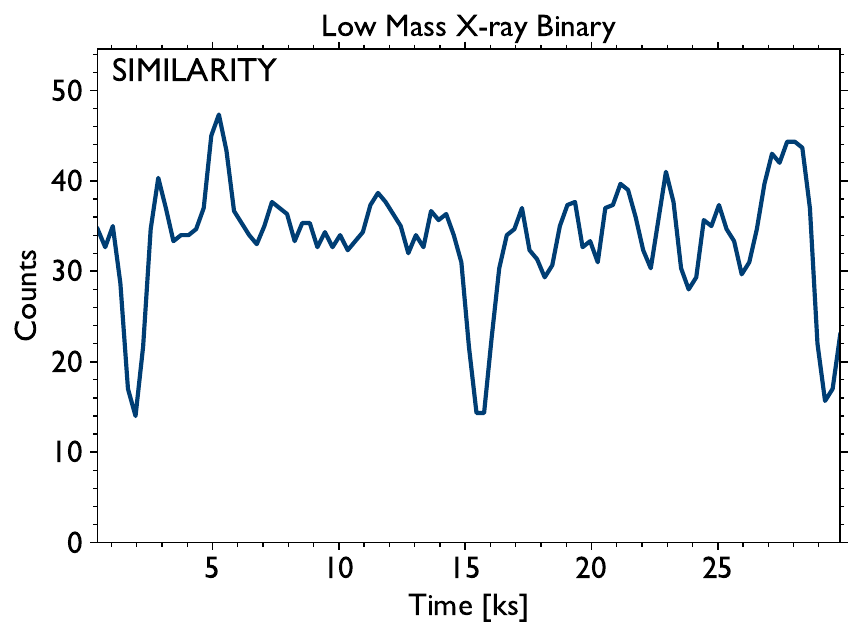}
\includegraphics[width=0.245\linewidth]{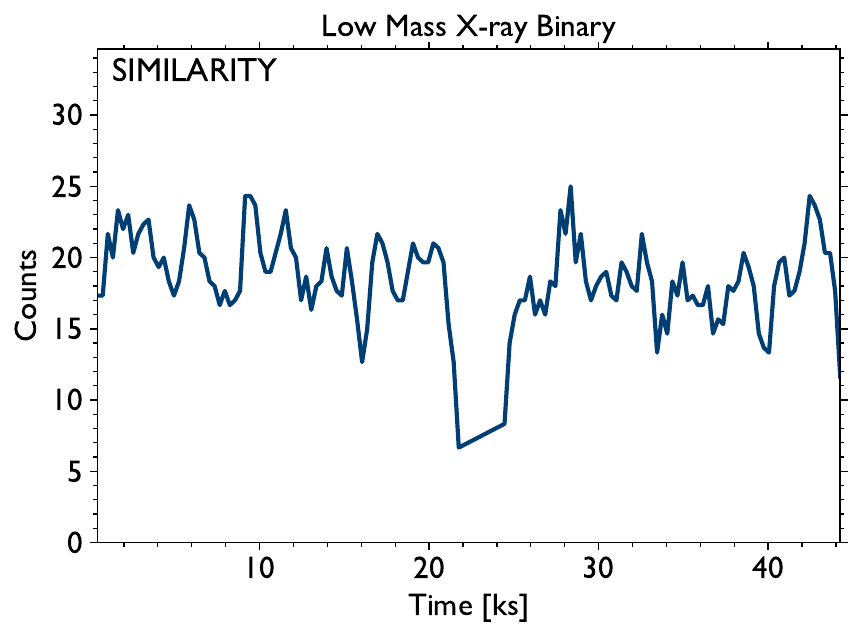}

\includegraphics[width=0.245\linewidth]{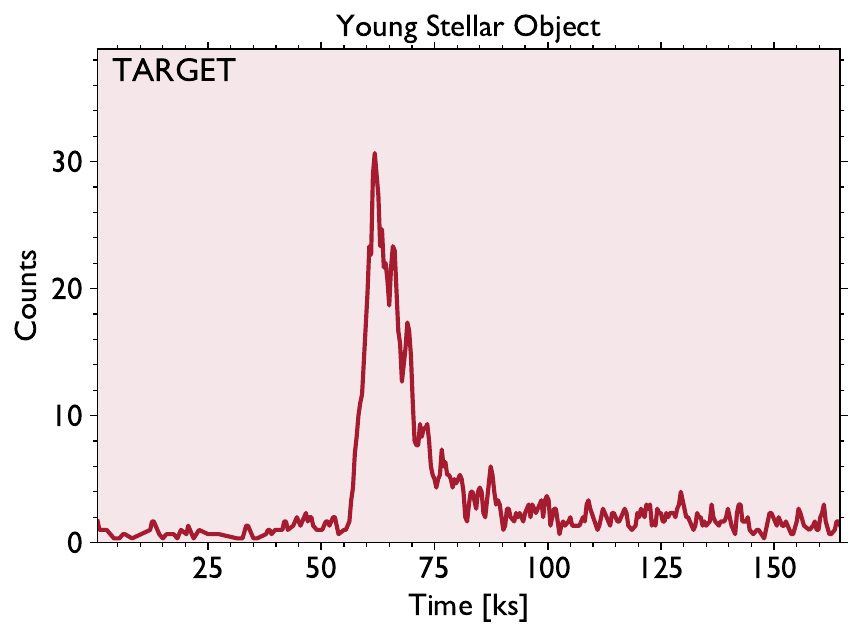}
\includegraphics[width=0.245\linewidth]{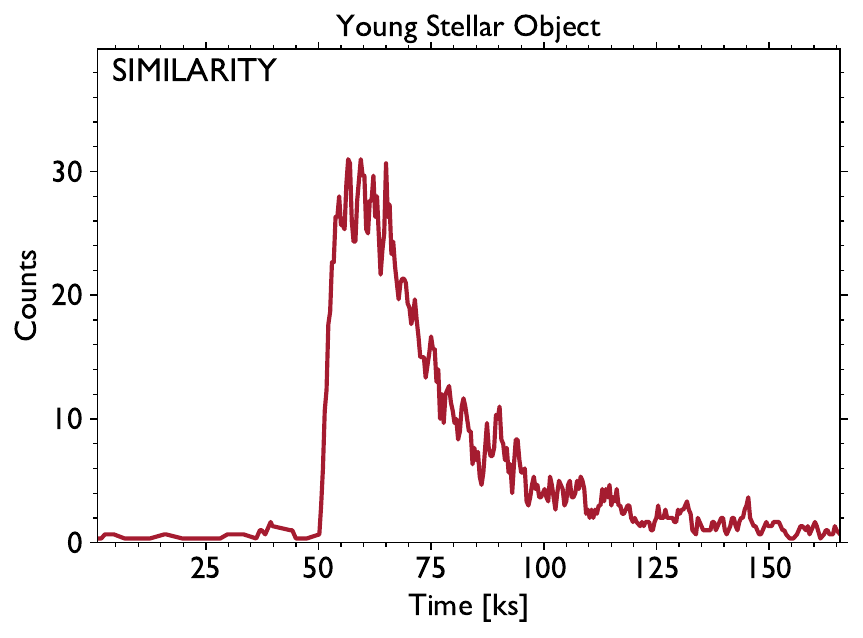}
\includegraphics[width=0.245\linewidth]{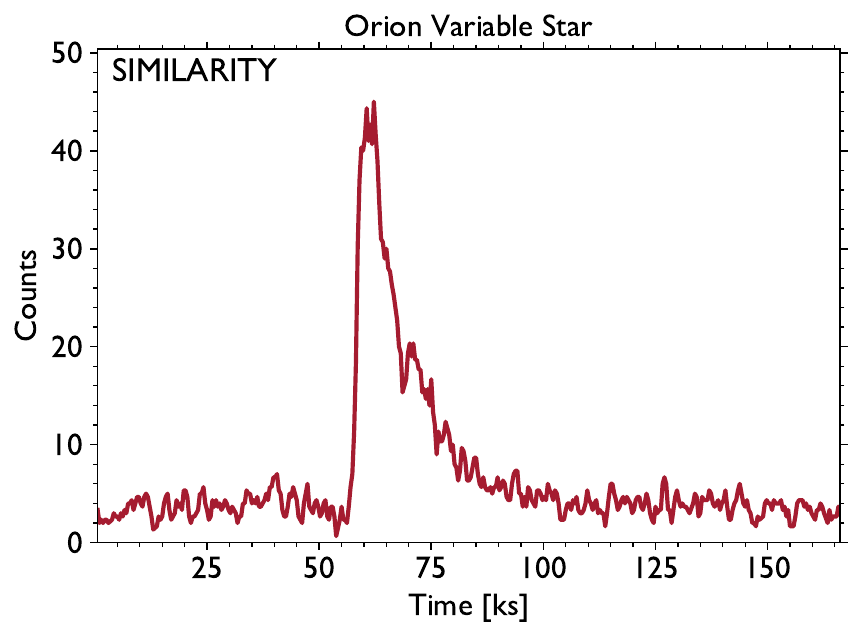}
\includegraphics[width=0.245\linewidth]{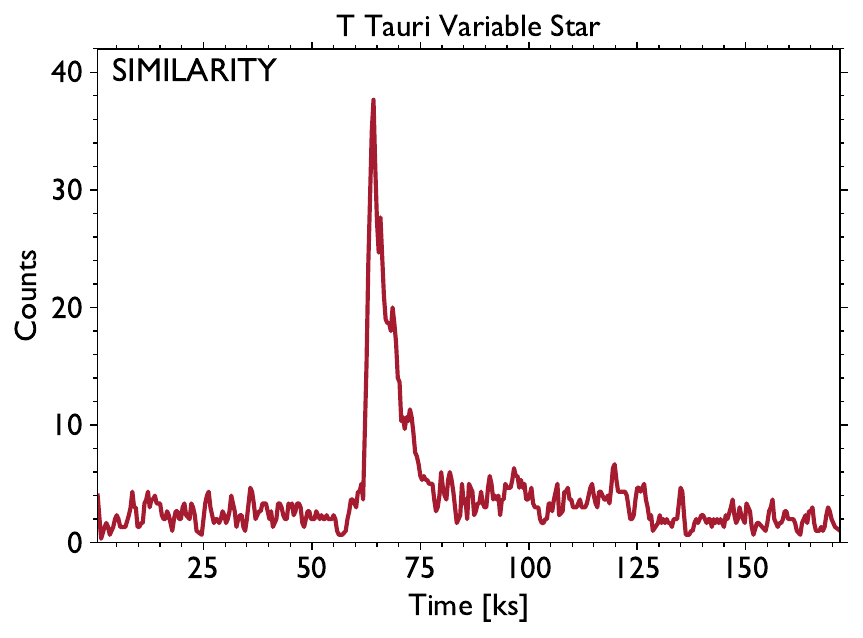}

\includegraphics[width=0.245\linewidth]{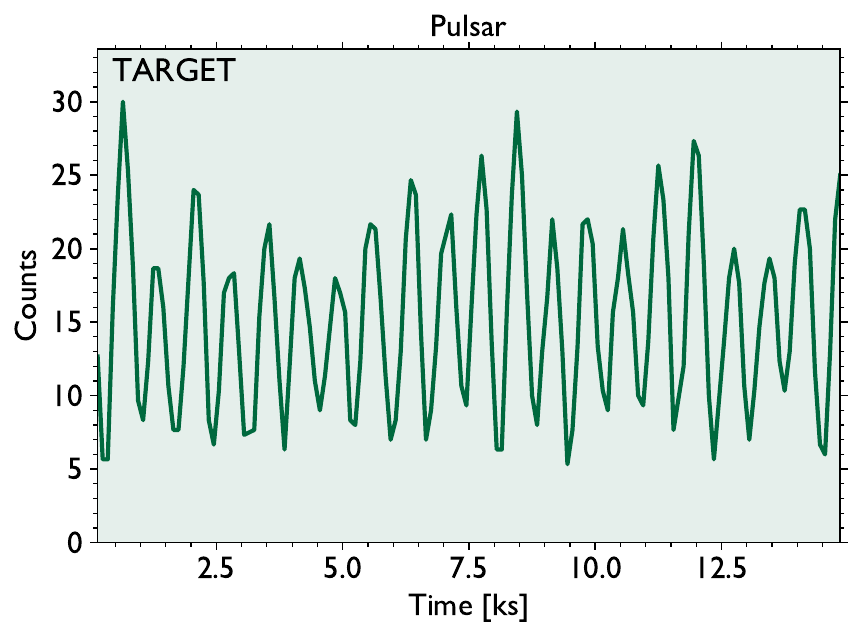}
\includegraphics[width=0.245\linewidth]{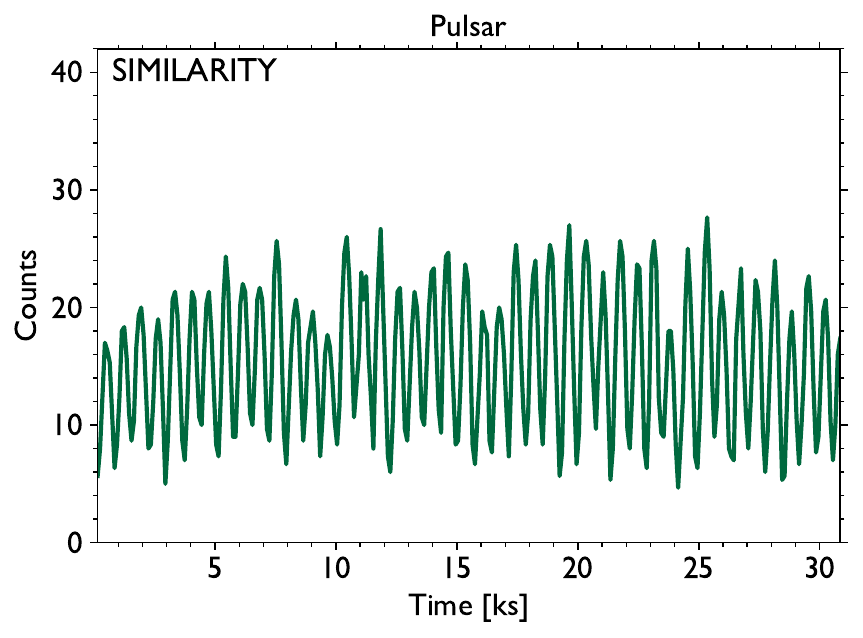}
\includegraphics[width=0.245\linewidth]{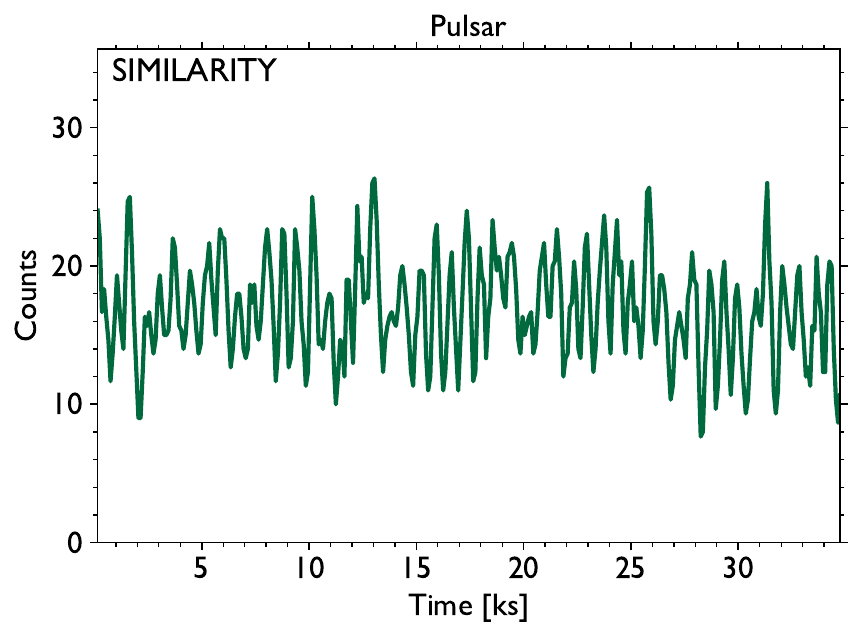}
\includegraphics[width=0.245\linewidth]{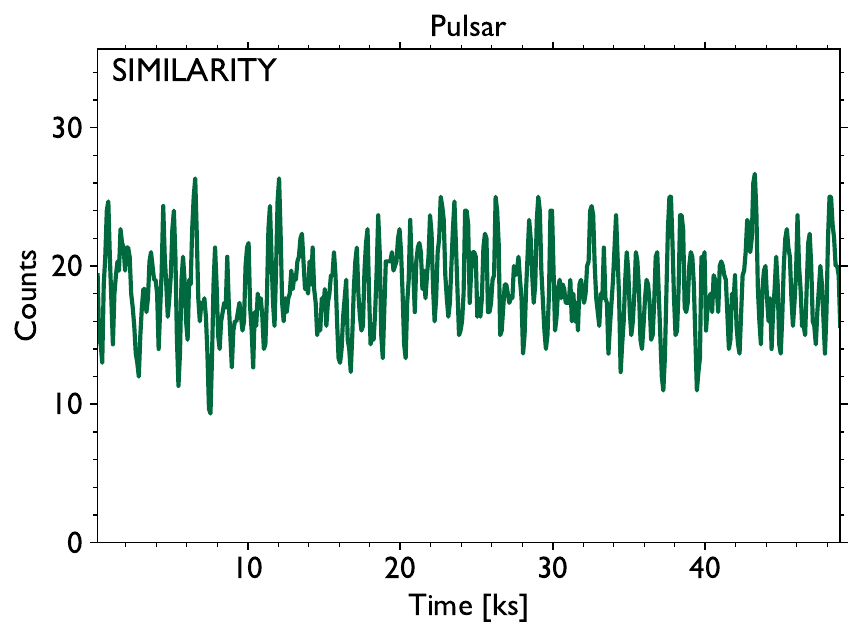}
        \caption{Nearest-neighbor retrieval results in the learned latent space for three representative transient types: dips from a low-mass X-ray binary (top row, 300~s bins), a flare from a young stellar object (middle row, 400~s bins), and pulsations from a pulsar (bottom row, 100~s bins). For each target light curve (column 1), we show its three nearest neighbors (columns 2–4). The retrieved neighbors correspond to physically similar phenomena: dips from low-mass X-ray binaries, flares from young stars or variable stars, and pulsations from pulsars.}
        \label{fig:anomaly}
\end{figure*}

\paragraph{Dimensionality Reduction \& Embedding Space} We visualize the learned representations as a two-dimensional embedding space using t-SNE \cite{van2008visualizing}. Figure~\ref{fig:tsne}A shows the embeddings color-coded by the CSC 2.1 variability index, which quantifies confidence in time variability from 0 to 10, with values above 6 indicating variability at a significance level of at least 2$\sigma$. The learned representations capture the temporal structure of the event files and effectively isolate highly variable X-ray sources, as evidenced by the well-isolated and pure dark cluster in the top-left embedding region. Figure~\ref{fig:tsne}B color-codes the embeddings by the hard-to-soft hardness ratio in CSC 2.1, which compares the relative fraction of photons detected in the soft (0.5–1.2~keV) and hard (2–7~keV) energy bands. Values near 1 indicate a hard spectrum; values near –1 indicate a soft spectrum. This metric is thus directly tied to the source's spectral properties. The smooth within-cluster hardness ratio gradients further demonstrate the model's ability to encode spectral information, allowing for queries of sources with distinct spectral properties. Figure~\ref{fig:tsne}C highlights the embedding positions of known flares (red), dips (blue), and pulsating sources (green). The dips include low-mass X-ray binaries, eclipsing binaries, and ultraluminous X-ray sources; the flares comprise extragalactic FXTs and young stellar flares; and the pulsations are predominantly from pulsars. In particular, each of these groups occupies well-separated regions of the embedding space. This clustering of physically meaningful source types confirms that the representations encode both spectral and temporal properties of the original event files. The embedding space of the SAE latents using the $E$–$t$–$dt$ cube inputs successfully separates different transient types into distinct clusters and further organizes them within clusters based on their spectral properties. This tendency to prioritize temporal behavior likely arises from the $E$–$t$–$dt$ cube representation having two temporal axes and only one for energy. The embedding space derived from the SAE using the $E$–$t$ maps as inputs is shown in Appendix \ref{appendix3}. Notably, in this case the known transients are dispersed all around the edges of the embedding space rather than forming distinct transient-rich clusters. Similarly, the embeddings in \citet{song2025poisson} do not show distinct transient source groupings. This underscores the added value of using a SAE on the $E$–$t$–$dt$ cubes for transient searches.

\paragraph{Semantic Clustering \& Unsupervised Classification}
By clustering the learned latent space using DBSCAN \cite{ester1996density}, we successfully isolate flare-dominant and dip-dominant groups, as shown in Figure~\ref{fig:tsne}C, leading to the identification of 3117 flare and 685 dip candidates. This yields a publicly available catalog of X-ray flares and dips \cite{dillmann_2025_14589318} that required only minimal manual filtering. Both clusters demonstrate high purity, with the flare cluster in particular containing a very high proportion of genuine flaring sources. Notably, we discover a previously unreported extragalactic FXT with unique temporal and spectral properties \cite{chandra_representation}, and a tidal disruption event (TDE) from a newly identified hyperluminous supersoft X-ray source \cite{sacchi2025}, both missed by previous transient searches in the Chandra archive \cite{2019MNRAS.487.4721Y, quirola2022extragalactic}. 

\paragraph{Anomaly Detection \& Similarity Search} Transient sources such as flares and dips represent only a small fraction of the dataset, which is largely dominated by relatively steady sources, making them anomalies from a data science perspective. The learned representations support anomaly detection and similarity-based retrieval of X-ray sources with distinct temporal and spectral features. In Figure~\ref{fig:anomaly}, we demonstrate this by retrieving the three nearest neighbors using $k$-nearest neighbors \cite{cover1967nearest} for three target sources: a dip from a low-mass X-ray binary, a flare from a young stellar object, and a pulsar. For each target source, the model retrieves similar light curves from the same or a closely related physical class, indicating that the latent space captures physically meaningful structure across source types.

\paragraph{Supervised Prediction Tasks} 
We train a vanilla XGBoost classifier and regressor \cite{chen2016xgboost} on the learned latent representations to distinguish highly variable X-ray sources (variability index $>$ 6) from non-variable ones (variability index $\leq$ 6), and to predict the hardness ratio of each source. The performances are summarized in Table~\ref{tab:performance_summary}. The classifier achieves a high variability classification accuracy of 0.97, outperforming the method proposed in \citet{song2025poisson}, and the hardness ratio regressor attains a slightly reduced R$^2$ score of 0.76, demonstrating that the learned representations encode temporal variability well and retain substantial predictive power for spectral hardness.
\begin{table}[t]
\centering
\caption{Performance summary of XGBoost models for variability classification and hardness ratio prediction. The models use 100 estimators and default hyperparameters without fine-tuning. An 80-20 train-test split is used for evaluation.}
\label{tab:performance_summary}
\vskip 0.08in
\begin{small}
\begin{tabular}{cll}
\toprule
\textsc{Task} & \textsc{Metric} & \textsc{Value} \\
\midrule
\multirow{3}{*}{\shortstack[c]{\textbf{Variability} \\ \textbf{Classification}}}
    & Accuracy                & 0.97 \\
    & F1-score (non-variable) & 0.98 \\
    & F1-score (variable)     & 0.72 \\
\midrule
\multirow{2}{*}{\shortstack[c]{\textbf{Hardness Ratio} \\ \textbf{Prediction}}}
    & R$^2$      & 0.76 \\
    & MSE                     & 0.03 \\
\bottomrule
\end{tabular}
\end{small}
\vskip -0.12in
\end{table}

\section{Conclusion}
We introduce a unified framework for learning sparse representations from event time series to enable a variety of different downstream tasks. We represent irregular event time series as $E$–$t$–$dt$ cube tensors and employ a SAE to learn sparse embeddings that capture physically meaningful variation while increasing their robustness to noise and other irrelevant features. The method requires minimal manual tuning and scales to large datasets across various domains, and serves as a foundation for subsequent analysis like anomaly detection, similarity search, and unsupervised classification. Applied to archival Chandra X-ray data, the learned representations encode temporal and spectral information and enabled the discovery of a new extragalactic FXT and a TDE from a hyperluminous supersoft X-ray source, both previously missed by other transient search methods. Future work involves optimizing the resolution of the $E$–$t$–$dt$ cubes, the deployment of more expressive SAE architectures and the application of the method to a variety of other scientific datasets.

\section*{Software and Data}

The data used in this paper was obtained from the publicly available CSC, using their public interfaces (\href{https://cxc.cfa.harvard.edu/csc/}{https://cxc.cfa.harvard.edu/csc/}). The preprocessed datasets are provided \href{https://drive.google.com/drive/folders/1LQNyM4in2RsOtEmxdh4u8viPY2fek9-3}{here}. All intermediate data products can be produced using the code provided in the \texttt{GitHub} repository \href{https://github.com/StevenDillmann/ml-xraytransients-mnras}{https://github.com/StevenDillmann/ml-xraytransients-mnras}. The catalog of transient candidates is publicly available on \href{https://zenodo.org/records/14589318}{Zenodo} and \href{https://cdsarc.cds.unistra.fr/viz-bin/cat/J/MNRAS/537/931}{VizieR}.

\section*{Acknowledgements}

This research has made use of data obtained from the Chandra Source Catalog provided by the Chandra X-ray Center (CXC). Juan Rafael Martínez-Galarza acknowledges support to AstroAI. We thank the reviewers for their insightful comments.

\section*{Impact Statement}

This paper presents work whose goal is to advance the field of Machine Learning and its applications to Astrophysics. There are many potential societal consequences of our work, none which we feel must be specifically highlighted here.



\bibliography{main_paper}
\bibliographystyle{icml2025}

\newpage
\appendix
\onecolumn
\section{$E$-$t$ Maps and $E$-$t$-$dt$ Cubes Visualizations} \label{appendix1}

\begin{figure*}[h!]  
    \centering
\includegraphics[width=0.275\linewidth]{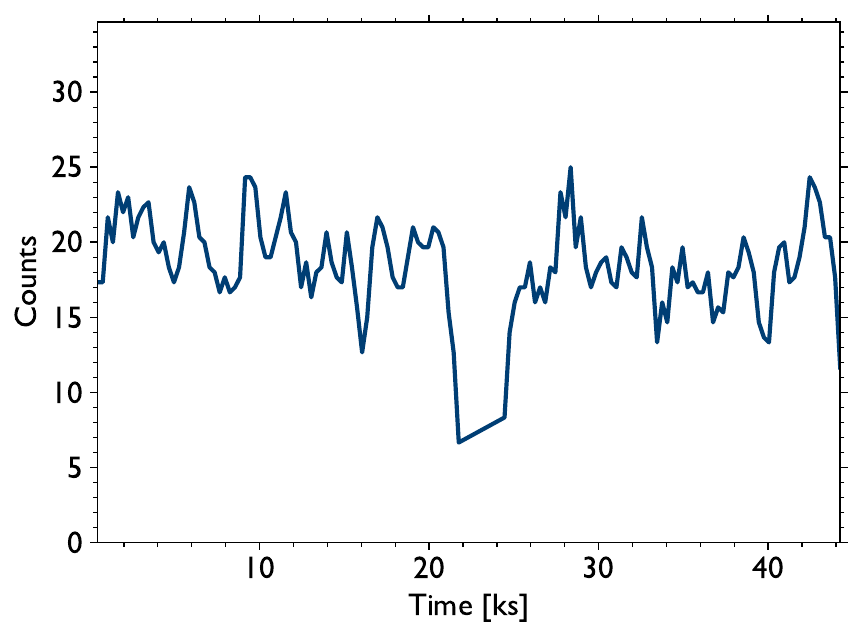}
\includegraphics[width=0.275\linewidth]{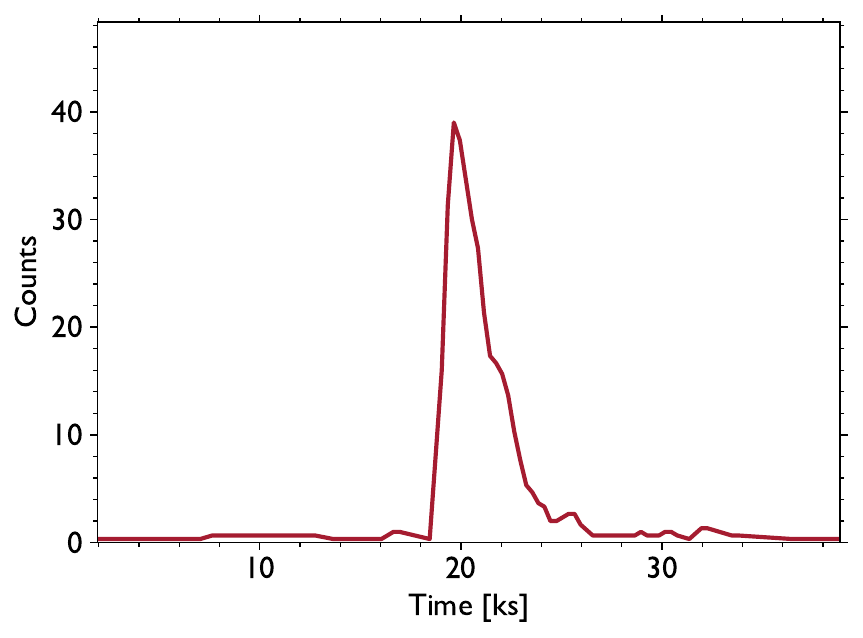}
\includegraphics[width=0.275\linewidth]{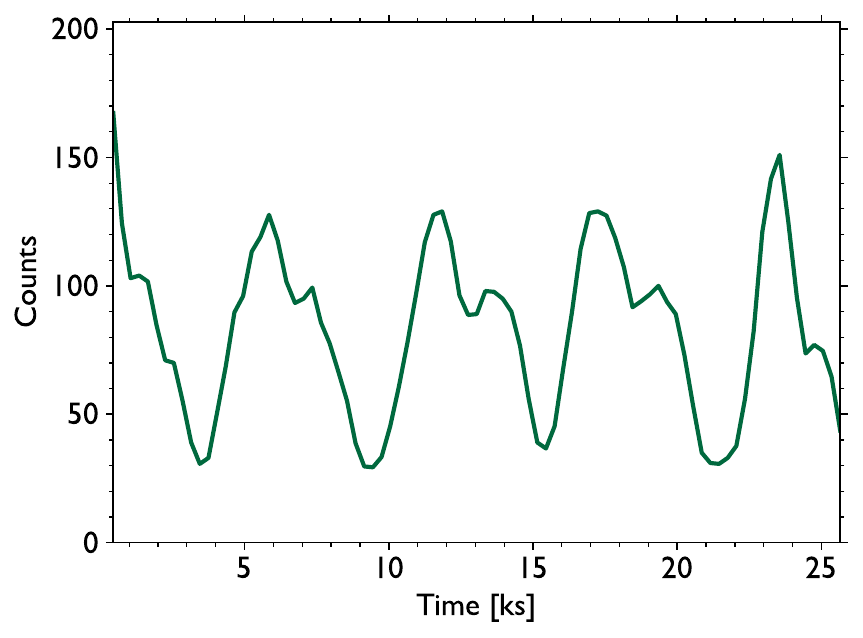}

\includegraphics[width=0.275\linewidth]{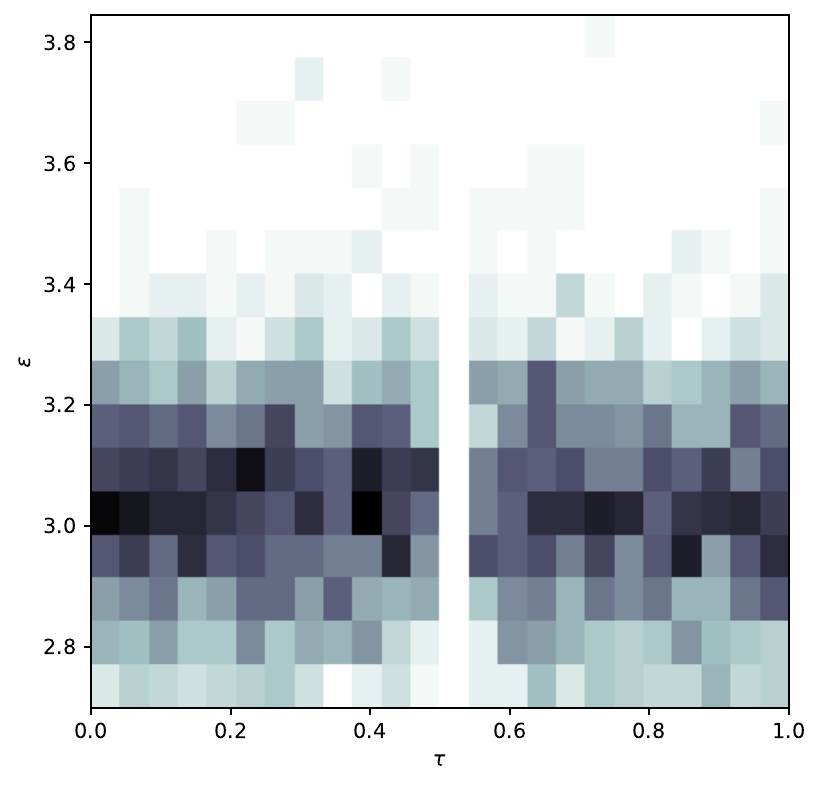}
\includegraphics[width=0.275\linewidth]{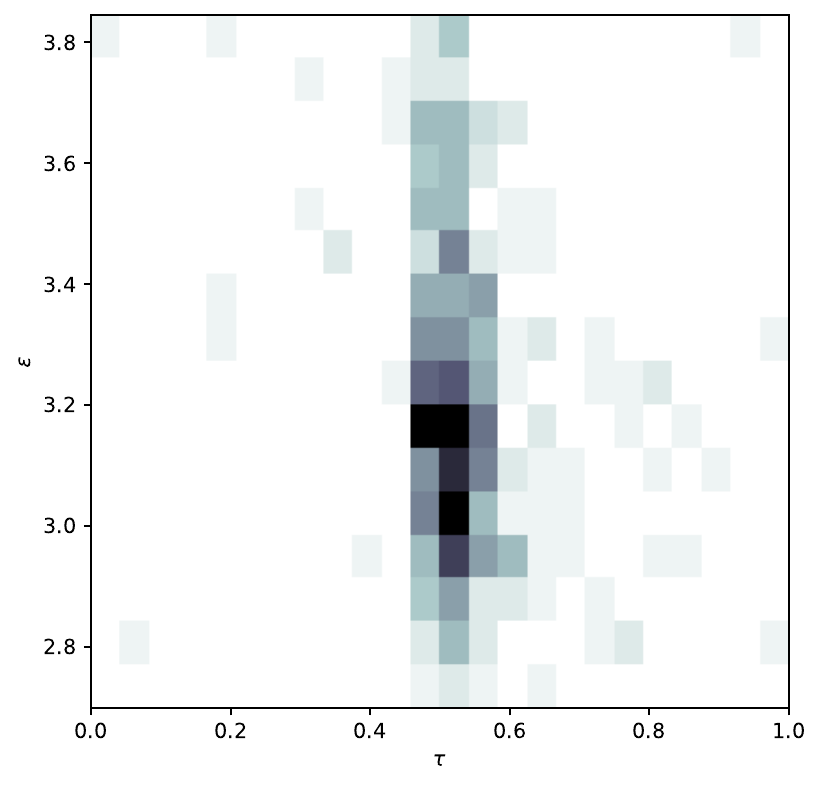}
\includegraphics[width=0.275\linewidth]{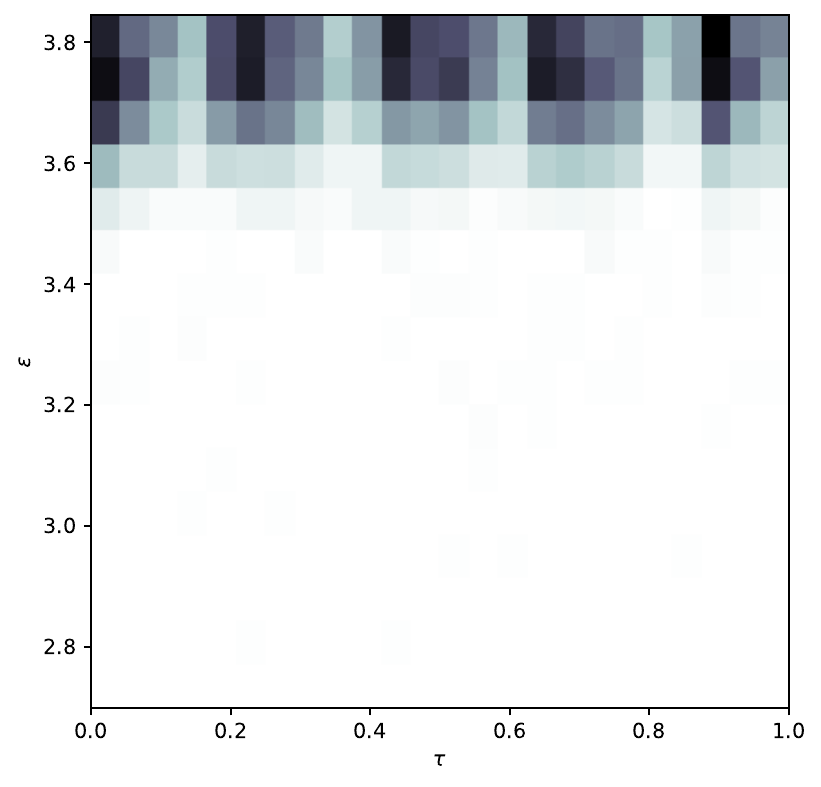}

\includegraphics[width=0.275\linewidth]{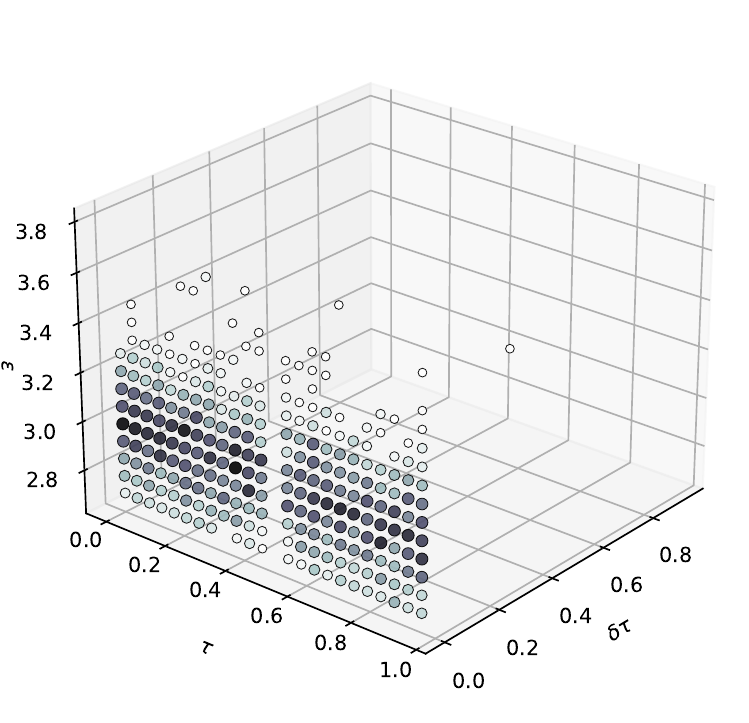}
\includegraphics[width=0.275\linewidth]{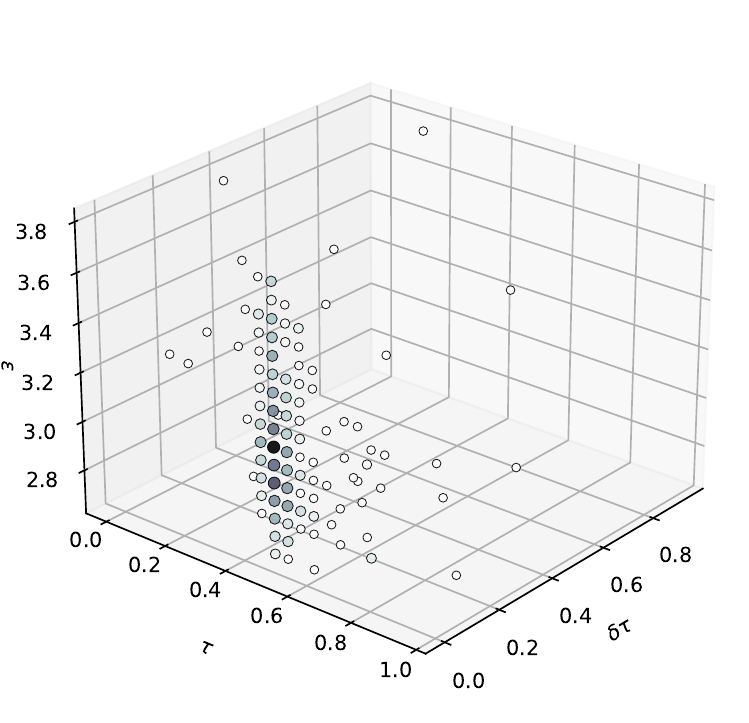}
\includegraphics[width=0.275\linewidth]{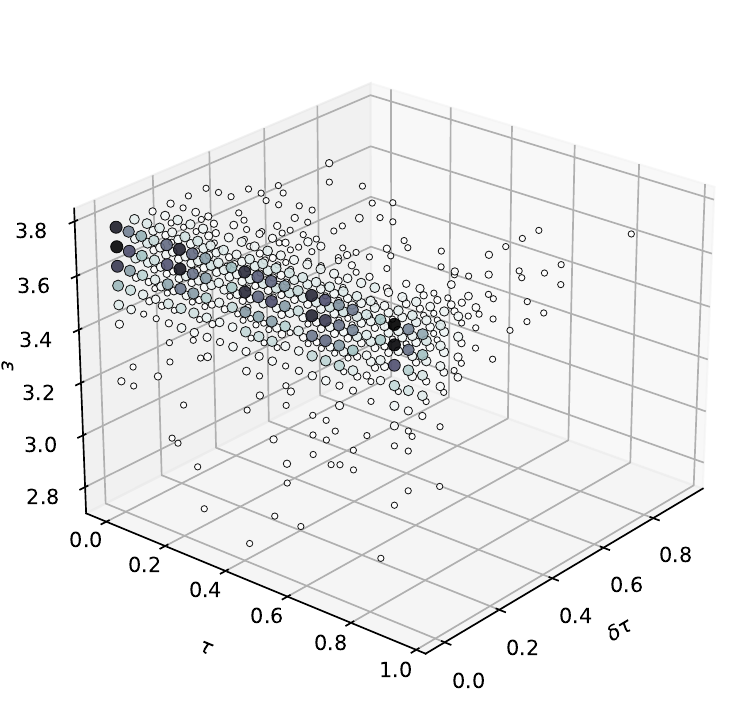}
        \caption{The top row shows 300~s bin light curves from event files featuring a dip (blue), a flare (red), and pulsations (green), respectively. The middle row shows the corresponding $E$–$t$ maps, and the bottom row shows the corresponding $E$–$t$–$dt$ cubes for these event files.}        
        \label{fig:mapscubes}
\end{figure*}

Figure~\ref{fig:mapscubes} displays the light curves, $E$–$t$ maps, and $E$–$t$–$dt$ cubes for event files featuring a dip, a flare, and pulsations, respectively.

\newpage
\section{SAE Architecture and Training} \label{appendix2}

The architecture of the experimental SAE for the $E$–$t$–$dt$ cube inputs is summarized in Table~\ref{tab:autoencoder2}. For the experiments presented in this paper, we flatten the input and use fully-connected layers for computational reasons. We recommend experimenting with more sophisticated SAE architectures to fully leverage the value of the $E$–$t$–$dt$ cubes and the presented SAE representation learning approach. Starting with the original dataset of 95,473 samples, we first apply a 90/10 split to obtain 85,925 samples for training and validation, and 9,548 for testing. We then further split the training and validation subset using an 80/20 ratio, resulting in 68,740 training samples and 17,185 validation samples. The model is trained for up to 200 epochs with a batch size of 1024. Optimization is performed using Adam \citep{kingma2014adam} with an initial learning rate of $0.01$. A learning rate scheduler reduces the rate by a factor of 10 if the validation loss plateaus for more than 10 epochs. We also apply early stopping based on the validation loss, which halts training if no improvement is observed for 25 consecutive epochs and restores the best-performing model weights. A similar procedure is followed for the experimental SAE with convolution layers summarized in Table~\ref{tab:autoencoder3} that we used for the $E$–$t$ map inputs.

\begin{table}[h!]
\centering
\caption{Summary of the encoder architecture of the fully-connected autoencoder used to extract informative features from the $E$-$t$-$dt$ cubes. Each layer uses Leaky ReLU activation and each standard fully-connected layer is followed by batch normalization with momentum 0.9.}
\vskip 0.08in
\begin{tabular}{ll}
\toprule
Layer & Output Shape \\
\midrule
Input & (24, 16, 16)  \\
Flatten & 6144  \\
Dense & 1536  \\
Dense & 384  \\
Dense & 92  \\
Dense (Bottleneck) & 24 \\
\bottomrule
\end{tabular}
\label{tab:autoencoder2}
\end{table}

\begin{table}[h!]
\centering
\caption{Summary of the encoder architecture of the convolutional autoencoder used to extract informative features from the $E$-$t$ maps. Each layer uses Leaky ReLU activation and each standard fully-connected layer is followed by batch normalization with momentum 0.9.}
\vskip 0.08in
\begin{tabular}{lllll}
\toprule
Layer & Output Shape & Filters & Kernel & Stride\\
\midrule
Input & (24, 16) & - & - & - \\
Convolution & (24, 16) & 32 & (3, 3) & - \\
Convolution & (12, 8) & 32 & (2, 2) & 2 \\
Convolution & (12, 8) & 16 &(3, 3) & - \\
Convolution & (6, 4) & 16 & (2, 2) & 2 \\
Flatten & 384 & - & - & - \\
Dense & 192 & - & - & - \\
Dense & 48 & - & - & - \\
Dense (Bottleneck) & 12 & - & - & - \\
\bottomrule
\end{tabular}
\label{tab:autoencoder3}
\end{table}

\newpage

\section{Embedding Space from the SAE applied on the $E$–$t$ Maps} \label{appendix3}

\begin{figure*}[h!]  
    \centering    \includegraphics[width=1\linewidth]{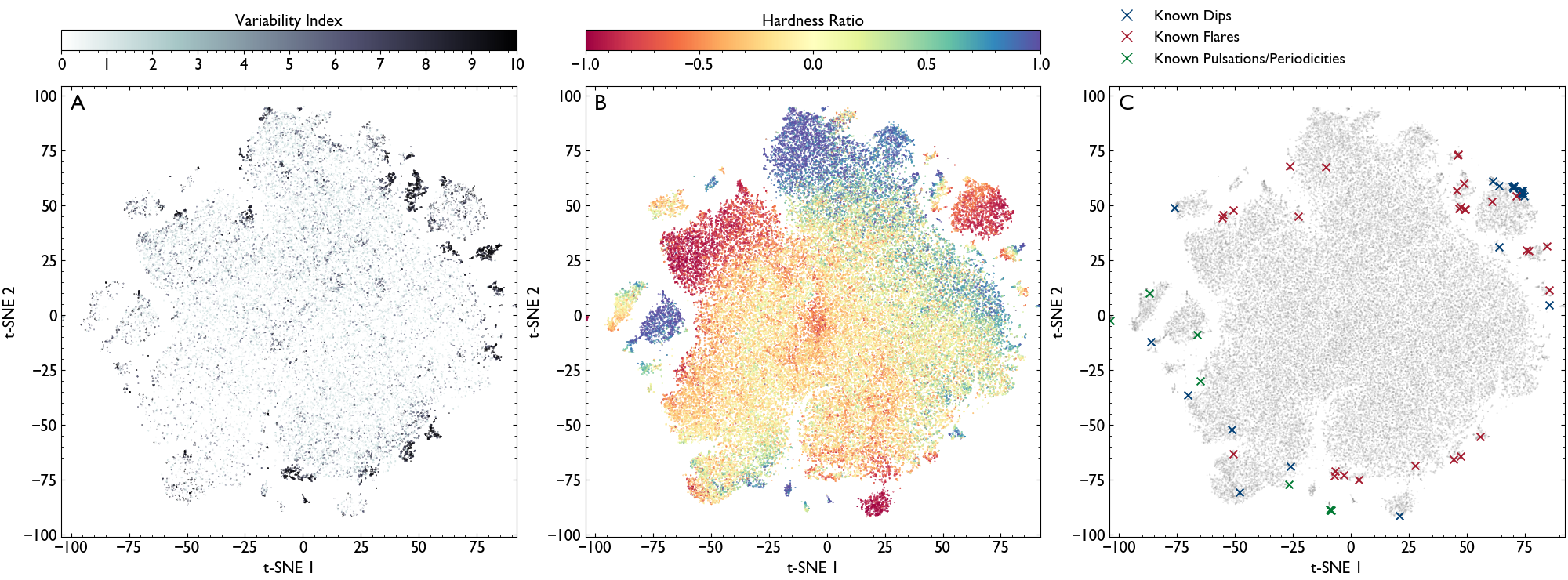}
    \caption{Two-dimensional t-SNE projection of the learned latent space from the SAE applied on the $E$–$t$ maps. Panel A: Points are color-coded by the variability index of the corresponding X-ray sources. Panel B: Points are color-coded by the hard-to-soft X-ray hardness ratio. Panel C: Known dips, flares, and pulsating sources (crosses) distributed across the embedding space.}
    \label{fig:tsne2}
\end{figure*}

Figure~\ref{fig:tsne2} shows the embedding space obtained from the experimental SAE applied on the $E$–$t$ maps.

\end{document}